\documentclass[11pt, letterpaper, logo]{googledeepmind}

\pdftrailerid{redacted}

\makeatletter
\renewcommand\bibentry[1]{\nocite{#1}{\frenchspacing\@nameuse{BR@r@#1\@extra@b@citeb}}}
\makeatother

\usepackage{kantlipsum, lipsum}
\usepackage{dsfont}
\usepackage{gdm-colors}
\usepackage{subfigure}
\usepackage{bbm}
\usepackage{wrapfig}
\usepackage{colortbl}
\usepackage{xcolor}
\usepackage{graphicx}
\usepackage{caption}
\usepackage[percent]{overpic}

\usepackage[most,skins,theorems]{tcolorbox}

\tcbset{
  aibox/.style={
    width=\linewidth,
    top=8pt,
    bottom=4pt,
    colback=blue!6!white,
    colframe=black,
    colbacktitle=black,
    enhanced,
    center,
    attach boxed title to top left={yshift=-0.1in,xshift=0.15in},
    boxed title style={boxrule=0pt,colframe=white,},
  }
}
\newtcolorbox{AIbox}[2][]{aibox,title=#2,#1}
\definecolor{lightblue}{rgb}{0.22,0.45,0.70}%

\newcommand{\eg}{e.g.\@\xspace}
\newcommand{\ie}{i.e.\@\xspace}

\PassOptionsToPackage{comma,numbers,sort,compress}{natbib}

\usepackage[comma,numbers,sort,compress]{natbib}

\usepackage{hyperref}[citecolor=magenta]

\hypersetup{
    colorlinks = true,
    citecolor = {magenta},
}

\usepackage{booktabs}
\usepackage{multirow}
\usepackage{array}
\usepackage[normalem]{ulem}
\usepackage{xspace}

\newcolumntype{G}{>{\color{gray}}c}

\usepackage{xcolor}  %
\usepackage{pifont}  %
\newcommand{\cmark}{\textcolor{green}{\ding{51}}}%
\newcommand{\xmark}{\textcolor{red}{\ding{55}}}

\newcommand{\supp}{Appendix\xspace}

\newcommand\mypara[1]{\vspace{0.5em}\noindent\textbf{#1.}}

\newcommand{\ourtitle}{
    BlenderFusion: 3D-Grounded Visual Editing and Generative Compositing
}

\newcommand{\ourmethod}{BlenderFusion\xspace}

\usepackage{arydshln}

\newcommand{\image}{{I}}
\newcommand{\camera}{{P}}
\newcommand{\render}{{R}}

\newcommand{\scene}{{S}}
\newcommand{\bbox}{{B}}
\newcommand{\source}{\text{src}}
\newcommand{\target}{\text{tgt}}

\graphicspath{{figures/}}

\title{\ourtitle}

\correspondingauthor{jca348@sfu.ca, shwoo@google.com}

\author[$\vardiamondsuit$,1,2]{Jiacheng Chen}
\author[1]{Ramin Mehran}
\author[1]{Xuhui Jia}
\author[1,3]{Saining Xie}
\author[1]{Sanghyun Woo}

\affil[1]{Google DeepMind}
\affil[2]{Simon Fraser University}
\affil[3]{New York University}
\affil[$\vardiamondsuit$]{Work done during an internship at Google Deepmind}
\vspace{-0.5cm}

\begin{document}

\begin{abstract}
We present \ourmethod, a generative visual compositing framework that synthesizes new scenes by recomposing objects, camera, and background. It follows a layering-editing-compositing pipeline: (i) segmenting and converting visual inputs into editable 3D entities (layering), (ii) editing them in Blender with 3D-grounded control (editing), and (iii) fusing them into a coherent scene using a generative compositor (compositing).
Our generative compositor extends a pre-trained diffusion model to process both the original (source) and edited (target) scenes in parallel. It is fine-tuned on video frames with two key training strategies:
(i) source masking, enabling flexible modifications like background replacement;
(ii) simulated object jittering, facilitating disentangled control over objects and camera.
\ourmethod significantly outperforms prior methods in complex compositional scene editing tasks. See the project page for demos and more results: \href{https://blenderfusion.github.io}{blenderfusion.github.io}
\end{abstract}

\maketitle

\section{Introduction}
\label{sec:intro}

\begin{figure*}[h]{}
\centering
\includegraphics[width=\linewidth]{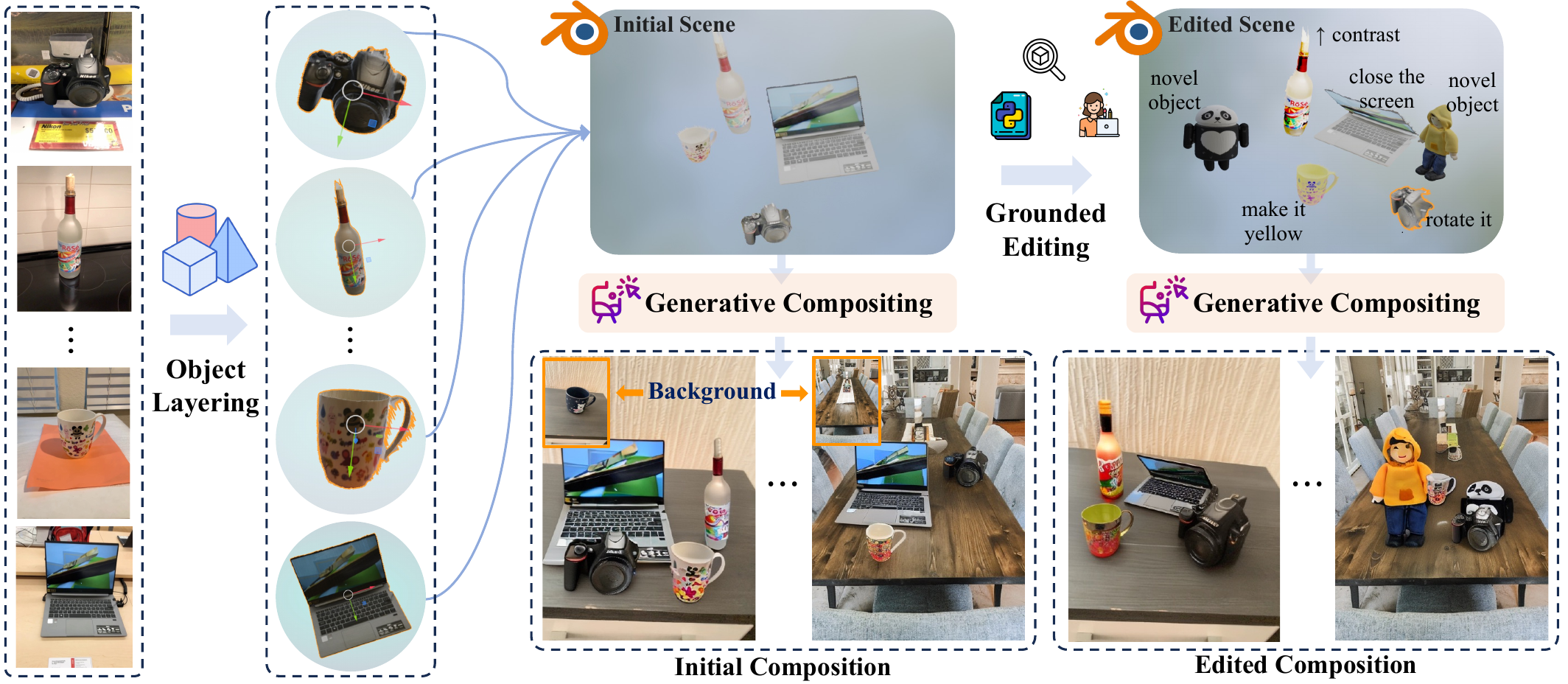}
\caption{
\ourmethod integrates the 3D-grounded editing capabilities of graphics software into the strong synthesis abilities of diffusion models. 
Despite fine-tuned only on video frames of simple object transformations with entangled camera motion, it learns precise object control, inherits Blender's rich editing functionalities (\eg, attribute modification, deformation, novel asset insertion), and generalizes to highly fine-grained multi-object editing and scene composition tasks (\autoref{fig:finegrained_control}).
}
\label{fig:teaser}
\end{figure*}

\textit{\textbf{Visual compositing}} is the process of constructing a scene by extracting objects from multiple images, manipulating their appearance or spatial configuration, inserting them into a new background, and adjusting the camera to produce a cohesive image or video. This enables novel visual narratives with high flexibility. While recent generative AI techniques excel at photorealistic text-to-image synthesis~\cite{karras2020style-gan,rombach2022latent-dm,saharia2022imagen,ramesh2022dalle-2,baldridge2024imagen-3}, they often fall short in complex compositing scenarios that demand precise, 3D-aware control over multiple scene elements, such as repositioning objects, modifying geometry and appearance, and adjusting viewpoint consistently.

To better characterize visual compositing tasks, we consider two key aspects: (i) the editable visual elements—objects, camera, and background—and (ii) the granularity of object-level control, including multi-object editing, novel object insertion, attribute modification, and non-rigid transformations. Some recent approaches blend objects from multiple input images while preserving identity and following coarse geometric layouts~\cite{hu2024instruct-imagen,song2022objectstitch}, but their control remains implicit, lacking 3D awareness and disentanglement. A more explicit line of work incorporates 3D-aware mechanisms to enhance editing fidelity and controllability.

Object 3DIT~\cite{michel2024object-3dit} enables text-driven 3D-aware edits via synthetic training data but is limited to rigid transformations of single objects. Neural Assets~\cite{wu2024neural-assets} disentangles object and background tokens for multi-object composition and spatial manipulation, yet lacks fine-grained control. Image Sculpting~\cite{yenphraphai2024image-sculpting} leverages Blender for accurate 3D edits but requires per-scene optimization and is mostly limited to editing a single object from a single input image. \autoref{tab:setting_comaprison} summarizes these representative methods across the two aspects above, highlighting that none achieves full-scene visual compositing with fine-grained, disentangled control over all core elements.

To address these limitations, we propose \ourmethod, a unified framework that mirrors the traditional visual compositing process through three fundamental steps (\autoref{fig:teaser}). (1) \emph{Layering}: off-the-shelf visual foundation models delineate foreground objects and lift them into 3D-editable entities. (2) \emph{Editing}: native Blender operations enable fine-grained, 3D-grounded modifications of object geometry, appearance, and camera viewpoint. (3) \emph{Compositing}: a diffusion-based visual compositor blends the Blender renderings with a background to produce a coherent image of the edited scene. By decoupling 3D control from image generation, \ourmethod combines the strengths of graphics-based editing and generative synthesis, enabling flexible, disentangled, and 3D-aware manipulation of objects, camera, and background.

\begin{table*}[t]
\centering
\scriptsize
\begin{tabular}{@{}lcccccccc@{}}
\toprule
\multirow{2}{*}{\begin{tabular}[c]{@{}c@{}}  Method \end{tabular}}  & \multicolumn{3}{c}{Visual Elements} & \multicolumn{4}{c}{Object Control}  & \multirow{2}{*}{\begin{tabular}[c]{@{}c@{}} Control \\ Interface \end{tabular}}  \\
\cmidrule(lr){2-4} \cmidrule(lr){5-8} 
 & Obj & Cam & BG & Multi-Obj & Novel-Obj & Attribute Change & Non-rigid Transform \\
\midrule
Object 3DIT~\cite{michel2024object-3dit} & \cmark & \xmark & \xmark & \xmark & \xmark & \xmark & \xmark & Text   \\
Neural Assets~\cite{wu2024neural-assets} & \cmark & \cmark & \cmark & \cmark & \xmark & \xmark  & \xmark  &  Obj tokens \\
Image Sculpting~\cite{yenphraphai2024image-sculpting} & \cmark & \xmark & \xmark & \xmark & \cmark & \cmark & \cmark  & Blender  \ \\
\midrule
\ourmethod (Ours) & \cmark & \cmark & \cmark & \cmark & \cmark & \cmark & \cmark  & Blender \\
\bottomrule
\end{tabular}
\caption{Capability comparisons between \ourmethod and existing 3D-aware editing works.}
\label{tab:setting_comaprison}
\vspace{-1em}
\end{table*}

Concretely, our diffusion compositor (\autoref{fig:pipeline}) operates on two parallel input streams: a source stream containing the original scene and a target stream reflecting the edited scene. To train the compositor effectively, we leverage object-centric videos and introduce two training strategies:
(1) \textit{Source masking}: To handle large contextual changes such as object insertion or removal, we mask the modified regions in the source stream, preventing them from interfering with target composition. At test time, this allows flexible masking to expose only the valid context.
(2) \textit{Simulated object jittering}: Since training videos are often dominated by camera motion, we introduce a reconstruction training setup that jitters object positions between source and target while keeping the camera fixed, thereby enriching supervision for disentangled object control.

We evaluate our method on three video datasets—MOVi-E~\cite{greff2022kubric}, Objectron~\cite{ahmadyan2021objectron}, and Waymo Open Dataset~\cite{sun2020waymo-open}. Although these datasets only exhibit basic object and camera motions, our method generalizes to diverse editing scenarios, enabling precise disentanglement of visual elements, fine-grained multi-object control, and complex compositional edits (\autoref{fig:teaser}, \autoref{fig:finegrained_control}).

\section{Related Work}
\label{sec:related}

\mypara{Visual Generation and Control}
Modern generative modeling, including Generative Adversarial Networks (GANs)~\cite{goodfellow2014gan,radford2015dc-gan} and diffusion models~\cite{sohl2015deep-nonequilibrium-thermodynamics,ho2020ddpm,song2020diffusion-sde}, first achieved high-fidelity content generation. 
The focus then shifted to user control, with natural language emerging as the primary interface~\cite{ramesh2022dalle-2,rombach2022latent-dm,saharia2022imagen}. 
However, text-based control is inherently limited.
To address this, specialized methods were developed for more granular control over aspects like geometry~\cite{zhang2023controlnet}, subject-driven generation~\cite{ruiz2023dreambooth,chen2022re,chen2023subject}, and aesthetics~\cite{sohn2023styledrop}.
Recent efforts have moved towards unifying these disparate controls into single, powerful models such as Instruct-Imagen~\cite{hu2024instruct-imagen}, which utilizes a versatile $ {text} + {image} \rightarrow {image} $ framework.
Despite these advancements, the reliance on text as the main control interface persists, along with its inherent limitations of ambiguity and difficulty in articulation. 
Our work addresses this by augmenting the generative model with a graphics engine. This approach leverages the powerful synthesis of diffusion models while gaining the precise control of a graphics engine, effectively disentangling the problems of generation and control.

\mypara{Visual Compositing}
Visual compositing—the process of assembling various visual elements into a single, seamless, and photorealistic image—is a crucial yet challenging task in visual generation.
Early diffusion-based methods focused on simple, single-object compositions guided by layouts~\cite{lu2023tf-icon,song2024imprint,chen2024anydoor,zhang2023controlcom,yuan2023customnet}.
While recent works have advanced to handle multiple objects~\cite{tarres2025multitwine} or even manipulate semantic attributes by mixing concepts from different images~\cite{garibi2025tokenverse}, these approaches remain largely confined to the 2D space. 
A comprehensive suite for complex, multi-object, and multi-image composition with precise geometric control remains an underexplored problem, which our work addresses.

\mypara{3D-aware Control}
A fundamental challenge in visual generation is achieving 3D-aware control that respects the physical nature of the world. Existing methods are primarily limited to single-object manipulation and often entangle visual elements, which prevents complex scene editing. 
Approaches such as GeoDiffuser~\cite{sajnani2025geodiffuser} and Diffusion Handle~\cite{pandey2024diffusion-handles} use 3D priors to transform model activations, while Magic Fixup~\cite{alzayer2024magic-fixup} implicitly learns world physics from video frames.
More recently, 3D-Fixup~\cite{cheng20253dfixup} explicitly learn this transformation through 3D edits.
A key limitation, however, is that these methods are all designed for single-object control.
As noted in \autoref{tab:setting_comaprison}, NeuralAsset~\cite{wu2024neural-assets} is a notable previous effort that addresses multi-object control, including the background and camera, albeit with some limitations.
In contrast, our framework provides a more robust solution by moving editing into a 3D graphics engine. 
We lift 2D elements into a 3D space for precise manipulation, then reproject the scene to 2D for rendering with a diffusion model, enabling effective and flexible compositing.

\mypara{Procedural Generation}
Aside from generating visuals directly from a model, notable efforts have formulated visual generation as a structured, procedural process. This involves representing the visual as a structured format, such as a Python program or script, and then relying on a graphics engine for rendering.
Frameworks such as BlenderAlchemy~\cite{huang2024blenderalchemy,gu2025blendergym}, FirePlace~\cite{huang2025fireplace}, and SceneCraft~\cite{hu2024scenecraft} leverage pre-trained vision-language models (VLMs) to generate desirable Blender Python scripts based on the user's edit intent, which is expressed through language. These systems generate and execute code using Blender to produce the final image. In this overall process, to steer the VLM toward desirable outputs, they devise methods to incorporate external information to aid the language model or algorithm for iterative self-correction.
Our framework is also loosely connected to procedural generation in the sense that we programmatically edit the projected 3D representation from 2D images using Blender. However, instead of relying on a pretrained language model to interpret user intent, our method empowers users to directly and interactively manipulate the meshes as they desire. The resulting render is then passed to our diffusion model to generate the final, high-fidelity image.

\section{\ourmethod}

\begin{figure}[t!]
\centering
\includegraphics[width=\linewidth]{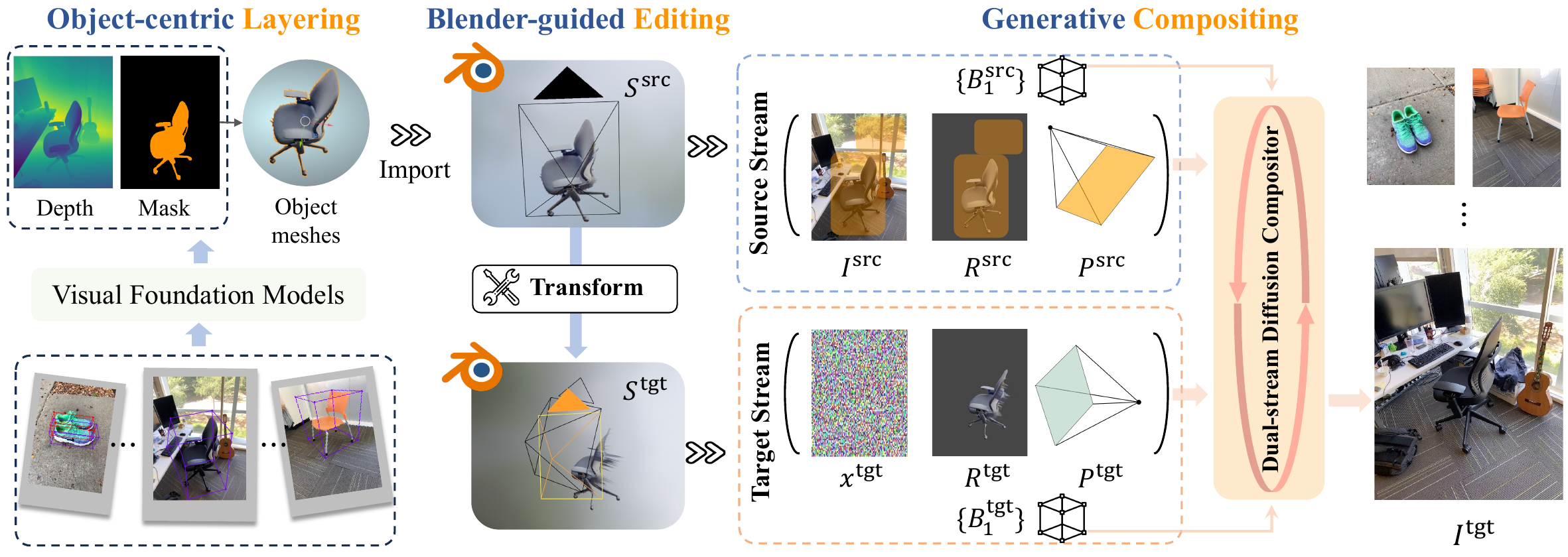}
\caption{\ourmethod employs a layering-editing-compositing process:
1) segment and lift objects from the source images into editable 3D elements; 
2) edit the visual elements in Blender to transform the initial scene to the target scene; and
3) use a dual-stream diffusion compositor to blend the target image. 
The video training data consists of object and camera pose annotations.
We use these annotations to simulate the test-time transformations in Blender.
The text input for each stream encodes a set of tuples consisting of object category labels and 3D bounding boxes.
The source masking strategy, indicated by overlaid \textcolor{orange}{orange} bounding boxes in $\image^\source$ and $\render^\source$, is detailed in \S\ref{method:refiner}. 
}
\label{fig:pipeline}
\end{figure}

The goal of this work is to enable precise, 3D-aware visual compositing by integrating generative models with a symbolic graphics tool, such as Blender, across the three fundamental steps of compositing: layering, editing, and compositing. Concretely, we obtain editable 3D object entities from multiple images (layering), import them into Blender for versatile 3D-grounded scene modifications (editing), and use a diffusion-based visual compositor to convert coarse Blender renderings and a background into realistic final images (compositing). \autoref{fig:pipeline} illustrates the full training pipeline.

A key component of the framework is the visual compositor, which refines the rendered target view to align with the intended edits. Since the reconstructed 3D scene $\scene^\source$ is derived from 2D images, its transformation often introduces noise, leading to artifacts in the target render $\render^\target$. The compositor corrects these artifacts with learned 3D shape priors to produce a coherent and photorealistic output. To train it effectively, we require paired scenes before and after edits. Object-centric videos naturally provide such supervision, but typically entangle object and camera motion, which limits generalization and disentangled control at test time. This motivates two simple yet effective training strategies, introduced in \S\ref{method:refiner}. We begin in \S\ref{method:pipeline} by detailing the full pipeline, including object segmentation and lifting, scene editing in Blender, and the architecture of the compositor model.

\subsection{\ourmethod Pipeline}
\label{method:pipeline}

\mypara{Object-centric 3D Layering}
The layering step leverages powerful off-the-shelf visual foundation models, such as SAM2~\cite{ravi2024sam2} and Depth Pro~\cite{bochkovskii2024depth-pro}, to obtain editable reconstructions of objects of interest from input images.
Our framework supports multiple input images (\autoref{fig:teaser}, \autoref{fig:finegrained_control}), and each image may contain multiple objects. For simplicity and generality, we assume that each object is reconstructed from a single image, while multiple views of the same object produce higher quality reconstructions, especially with recent advances in 3D foundation models~\cite{wang2024dust3r,leroy2024mast3r,wang2025vggt}.

We begin by projecting 3D bounding boxes into the image space to obtain coarse 2D boxes. However, since these projected 2D boxes are usually loose, we refine them using Grounding DINO~\cite{liu2024grounding-dino}, prompting it with object category labels. In practice, if the predicted box from grounding DINO has a high overlap with the projected one (IoU $>$ 0.5), we replace the latter.
With the refined 2D box, we then prompt SAM2 to extract the object mask. Combining this mask with metric depth predictions from Depth Pro, we obtain object-wise depth. We then adjust the scale of the predicted depth to align with each object's 3D bounding box.
Finally, we back-project adjusted object-wise depth maps to generate 3D point clouds and connect adjacent points to form a triangle mesh. This results in a set of 3D entities, $\scene^\source$, which are imported into Blender for subsequent editing.

Note that the layering process described above is efficient and applied to all the data. 
At test time, more advanced layering operations can be included for higher-quality reconstructions. 
When the editing task requires a complete and high-quality 3D model (\eg, complicated intra-object part-level editing, or material change, etc.), we optionally use state-of-the-art image-to-3D models~\cite{xiang2024trellis,zhao2025hunyuan3dv2} to produce complete object meshes. 
Concretely, we crop out the image patch of each object using the SAM2 object mask and run Hunyuan3D v2~\cite{zhao2025hunyuan3dv2} with the cropped image to obtain a complete textured mesh.
We then align the mesh with the object's 3D box and the 2.5D surface reconstruction.

\mypara{Blender-guided Editing}
The output of the layering step, $\scene^\source$, is then imported into Blender or other graphics software, where versatile transforms can be applied to objects and the camera with precise 3D grounding. \ourmethod covers the following control tasks:  

\noindent \raisebox{0.25ex}{\tiny$\bullet$} {\it \textbf{Basic object control}} includes the translation, rotation, or scaling of each independent object, as well as object removal, insertion, or replacement. Since $\scene^\source$ contains per-object 3D models, all these transformations can be applied automatically and reflected in the rendering $\render^\target$.

\noindent \raisebox{0.25ex}{\tiny$\bullet$} {\it \textbf{Advanced object control}} represents object attribute change (\eg, color, material), non-rigid object transforms (\eg, part-level control, deformation, etc.), and novel object insertion (\eg, not covered by the training data). These operations are inherited from Blender and can be fulfilled by user interactions in the user interface or automatic scripts (\autoref{fig:teaser} and \autoref{fig:more_results}).

\noindent \raisebox{0.25ex}{\tiny$\bullet$} {\it \textbf{Camera and Background control}} involves camera motion and background replacement. The new background is specified with an image to replace $\image^\source$, while the camera motion is simulated through the camera object in $\scene^\source$.

After applying all these edits, both the original and edited scenes, $\scene^\source$ and $\scene^\target$, are rendered into $\render^\source$ and $\render^\target$, providing reliable 3D-grounded control signals for the compositing step. $\render^\source$ and $\render^\target$ contain the rendered RGB image and an object index mask from Blender's Object Index Pass. 
Although the training data only covers simple object transformations and camera motion, our generative compositor generalizes to all the above editing tasks at test time.

\mypara{Generative Compositing}
The generative compositor has access to two streams of input information (see \autoref{fig:pipeline}). The {\it source stream} consists of the image $\image^\source$, its rendering $\render^\source$, camera parameters $\camera^\source$ and object poses $\bbox^\source$. 
The {\it target stream} includes the edited rendering $\render^\target$, its camera parameters $\camera^\target$, and object poses $\bbox^\target$. $\render^\target$ may be noisy due to imperfect reconstruction and transformation. At test time, both the source and target streams may contain objects extracted from multiple input images.

To effectively process the dual-stream input, we adapt a pre-trained diffusion model with three key architectural modifications based on
Stable Diffusion v2.1~\cite{rombach2022latent-dm}.
First, we extend the model to a dual-stream architecture, where a single weight-shared denoising UNet processes both streams independently while enabling interaction through self-attention~\cite{shi2023mvdream,tang2024mvdiffusion++,gao2024cat3d}. 
Second, we modify the first layer of the UNet to accommodate additional conditional inputs, increasing its channels from 4 to 15 with zero-initialized weights.
The original 4 channels process the VAE-encoded image or noise for the source or target stream. 5 additional channels handle Blender renderings (4 for the VAE-encoded rendering image and 1 for the instance mask). The remaining 6 channels encode camera parameters using Plücker embeddings~\cite{sitzmann2021light-field-network,gao2024cat3d}.
Third, each stream has an independent set of text tokens consisting of tuples of object class labels and poses. The labels are CLIP-embedded~\cite{radford2021clip} while 3D boxes are converted into positional encodings~\cite{vaswani2017transformers} and processed by an MLP. The resulting embeddings are concatenated into a sequence as the text tokens for their respective streams.

\subsection{Generative Compositor Training}
\label{method:refiner}

The overview of the generative compositor is in \autoref{fig:pipeline}.
To simulate test-time editing, we randomly sample two frames from a video: a source (original) and a target (edited).
$\image^\source$ goes through the layering step to obtain $\scene^\source$ and $\render^\source$. $\scene^\source$ is then transformed with the 3D object bounding boxes (object pose) and camera parameters (view changes) to obtain $\scene^\target$, and re-rendered into $\render^\target$. 
This process introduces noise into the target render within the reconstruction, alignment, and transformation process.
Given the noisy target render, the model is trained to complete missing object textures and geometry, filling in the view-changed background, using the source stream as context.
While effective, this training strategy has two key limitations.
First, the model struggles with edits that intensively modify the original context, such as object removal, insertion, or background change.
Second, it performs poorly with disentangled object control, especially when the camera remains fixed.
To this end, we introduce our two specific training strategies.

\mypara{Source Masking} 
If the original context is altered (\eg, object is removed, replaced, or aggressively edited), the model should disregard the modified region in the source when compositing the target image.
To achieve this, we introduce the source masking strategy during training, where each object in $\image^\source$ and $\render^\source$ is randomly masked out with a probability of 0.5 (see the overlaid \textcolor{orange}{orange} boxes in \autoref{fig:pipeline}). At test time, we then flexibly mask both the source image and source render, or only mask the source image, depending on the concrete control task.
This training strategy also has a regularization effect, which mitigates over-reliance on source information and the relative camera pose, and enforces the model to more accurately follow the intended edits in the target render (\autoref{fig:ablation}).
We also apply random masking to background regions to prevent the model from getting an inpainting bias. We refer to \S\ref{supp:method_details} for complete details.

\begin{wrapfigure}[13]{r}{0.5\textwidth}     \centering
\vspace{-1em}
\includegraphics[width=\linewidth]{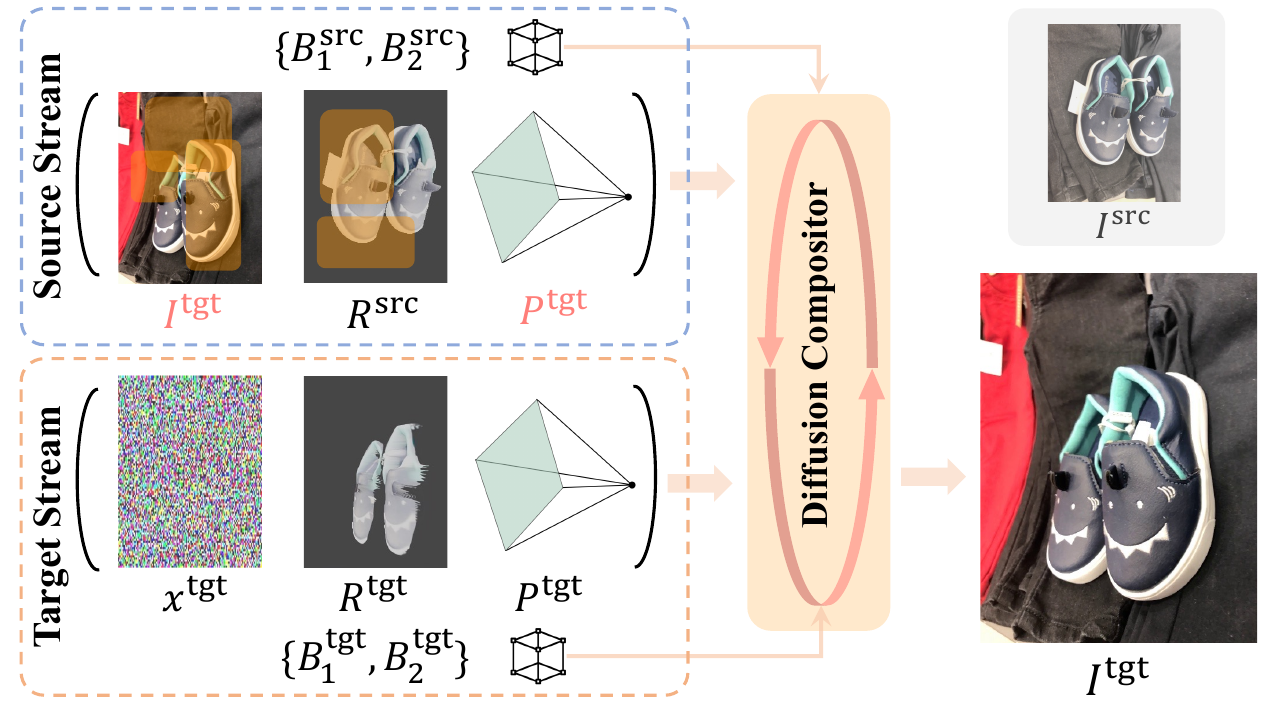}
\caption{The simulated object jittering training strategy for improving disentangled object control.}
\label{fig:recon_training}
\end{wrapfigure}

\mypara{Simulated Object Jittering}
We observe that object motion supervision in training videos is limited and often strongly entangled with camera motion. For example, in object-centric videos such as Objectron, objects frequently remain static while only the camera moves. To offset this issue, we introduce an object jittering training strategy that simulates dynamic object motions under a fixed-camera setup, as shown in \autoref{fig:recon_training}. The key change from the standard video learning setup is the replacement of $\image^\source$ and $\camera^\source$ with $\image^\target$ and $\camera^\target$. A random source masking strategy is applied separately to $\image^\target$ and $\render^\source$. To infer the masked $\image^\target$ from the noisy $\render^\target$, the model learns to effectively leverage the object information in $\render^\source$, $\bbox^\source$, and $\bbox^\target$, while the camera motion remains fixed. Despite its simplicity, this approach provides accurate, disentangled control of the object and camera at test time.

\section{Experiments}

In \S\ref{exp:setting}, we describe the implementation details and experimental setups.
We then provide both quantitative and qualitative evaluations in \S\ref{exp:standard_results} and \S\ref{exp:finegrained_results}, comparing \ourmethod with leading approaches on 3D-aware control tasks of increasing complexity and finer granularity. Finally, \S\ref{exp:more_results} presents more results on human evaluation, generalization beyond training data, and ablation studies.

\subsection{Experimental Settings}
\label{exp:setting}

We implement our generative compositor using the Diffusers framework~\cite{von-platen-etal-2022-diffusers} and finetune the publicly available Stable Diffusion v2.1 base model. The model is trained with 8 NVIDIA A100 80GB GPUs. We use the v-prediction diffusion training objective~\cite{salimans2022progressive-distill}. The batch size is 320, and the model is trained for 30,000 iterations. We turn on gradient checkpointing, use gradient accumulation of two steps, and train the model with mixed bfloat16 precision. We use AdamW optimizer with a weight decay factor 1e-2 and a 500-step linear warmup. The learning rate is 5e-5 for the diffusion model and 1e-4 for the new MLP that encodes the object 3D box. For inference, we run the DDPM~\cite{ho2020ddpm} sampler for 50 steps, and classifier-free guidance (CFG)~\cite{ho2022diffusion-cfg} is used with scale 2.0. When encoding the camera parameters, the camera of the source stream sets the reference coordinate frame. The 3D bounding box is projected into the image plane, and each corner is represented by (x, y, depth). 
The ratios of vanilla video training, training with source masking, training with both source masking and simulated object jittering, and unconditional training (for CFG) are 0.35, 0.3, 0.3, and 0.05, respectively. We then introduce datasets, baselines, and metrics. At test time, the layering step only uses the 2.5D surface reconstructions without running an image-to-3D model for complete meshes, except for the complex editing tasks in \autoref{fig:more_results} (Bottom).
Please refer to \S\ref{supp:details} for complete implementation details.

\mypara{Datasets}
We cover three public video datasets with 3D object and camera annotations:

\noindent \raisebox{0.3ex}{\tiny$\bullet$} {\it MOVi-E} is a synthetic multi-object video dataset from the Kubric~\cite{greff2022kubric} dataset generator. To obtain videos with more objects and camera dynamics, we run Kubric's official data generation pipeline to produce 10,000 MOVi-E videos by increasing the number of dynamic objects and the camera motion range. The image resolution is 512$\times$512. This MOVi-E variant provides a challenging benchmark for multi-object control with 3D awareness (occlusion, novel viewpoint, etc.).

\noindent \raisebox{0.3ex}{\tiny$\bullet$} {\it Objectron~\cite{ahmadyan2021objectron}} contains 15,000 real-world object-centric video clips across 9 categories, recorded with camera movement in real-world environments. 
Note that this dataset only contains static object. 
We keep the aspect ratio of the dataset and resize the images to 384$\times$512.  

\noindent \raisebox{0.3ex}{\tiny$\bullet$} {\it Waymo Open Dataset (WOD)~\cite{sun2020waymo-open}} consists of 1,000 real-world videos captured from self-driving cars. Following prior work~\cite{wu2024neural-assets}, we use the front-view camera and filter out small cars. 
We keep the aspect ratio of the original dataset and resize the images to 528$\times$352.

\mypara{Baselines} 
We use 3DIT and NA as baselines, while Image Sculpting is excluded as it is a per-scene optimization method and mainly handles single objects.
For controlled experiments between baselines and ours, we re-implement 3DIT and NA within the Diffusers framework, strictly following the descriptions in their original papers.
We ensure all approaches use the same base Stable Diffusion v2.1, same input information (input image, object category, source/target object 3D box), identical training setups (training loss, training iterations, learning rate, etc.), and identical sampling setups (sampler and sampler steps). As a result, the primary difference lies in their core controllability strategy.
Concretely, 3DIT directly uses the text embeddings (\ie, class labels and source/target 3D boxes) to transform the source image; 
NA employs external DINO~\cite{caron2021dino-v1} to extract the per-object appearance and combines it with target 3D box, then controls through the text embedding interface;
\ourmethod leverages a more explicit interface, Blender, to obtain source and target renders as 3D-grounded control signals, where many fine-grained 3D visual edits are possible.
We detail the baseline approaches and their changes below.

\noindent \raisebox{0.3ex}{\tiny$\bullet$} \emph{Object 3DIT}~\cite{michel2024object-3dit} originally adapts Zero-1-to-3~\cite{liu2023zero123} to incorporate a text instruction describing the control task and is trained on simple synthetic data, mainly targeting single-object control.
For a fair comparison, we introduce several updates.
The base model is replaced wit Stable Diffusion (SD) v2.1, and the source image is encoded by the SD VAE and concatenated with the input, following the original approach.
Additionally, the Plücker embedding of the relative camera pose is concatenated. 
Instead of using a plain text instruction, we replace it with serialized object embeddings to handle multi-object control.
Each object embedding consists of the CLIP embedding of its object category and the encoding of its 3D bounding box.

\noindent \raisebox{0.3ex}{\tiny$\bullet$} \emph{ Neural Assets (NA)}~\cite{wu2024neural-assets} enables multi-object 3D editing through object tokens that encode appearance and pose features. The object appearance is obtained by applying RoIAlign~\cite{he2017mask} to DINO features of the foreground image, while the pose is obtained from the 3D object bounding box with an MLP. The background is processed separately, with the appearance extracted from the foreground-masked DINO features and the pose computed from the relative camera pose between the source and target images. 
Tokens for all objects and the background are serialized into a sequence, replacing the text embedding in the pre-trained Stable Diffusion v2.1 architecture.

\mypara{Evaluation Metrics}
We follow the evaluation metrics used by NA~\cite{wu2024neural-assets}. 
PSNR, SSIM, and LPIPS are computed at both image and object levels.
The image-level metrics include FID, while the object-level metrics include the per-object DINO feature cosine similarity. Complete details are in the \supp. 
All quantitative evaluations follow the standard video frame-based setup, which measures how well the model transforms a source frame into a target frame given camera and object poses.
However, this setup is not ideal, especially for assessing fine-grained, disentangled control tasks, which are more relevant and practical in real-world editing scenarios but do not have ground truth answers.
Thus, we rely on qualitative comparisons and human evaluations to better evaluate these capabilities.

\mypara{Synthetic Data Pre-training}
We observe that all approaches struggle to learn disentangled object rotation on WOD, primarily due to the lack of angular object motion in the dataset (\ie, most cars exhibit only simple translation across frames). To address this, we initialize WOD training with the pre-trained checkpoint from MOVi-E, which contains richer object motions and provides stronger 3D object shape priors. For all three approaches, we adopt this initialization and reduce the base learning rate to $1\mathrm{e}{-5}$. More analyses are in ~S\ref{supp:failure}.

\begin{table}[!t]
\small
\centering
\resizebox{0.95\columnwidth}{!}
{
\begin{tabular}{llcccccccc}
\toprule
\multirow{2}{*}{\begin{tabular}[c]{@{}c@{}} Dataset \end{tabular}} & \multirow{2}{*}{\begin{tabular}[c]{@{}c@{}}  Model \end{tabular}} & \multicolumn{4}{c}{\textbf{Object-level metrics}} & \multicolumn{4}{c}{\textbf{Frame-level metrics}}  \\ 
\cmidrule(lr){3-6} \cmidrule(lr){7-10}
&  & PSNR $\uparrow$ & SSIM $\uparrow$ & LPIPS $\downarrow$ &  DINO $\uparrow$ & PSNR $\uparrow$ & SSIM $\uparrow$ & LPIPS $\downarrow$ & FID $\downarrow$ \\ 
\midrule

\multirow{3}*{\shortstack[c]{MOVi-E~\cite{greff2022kubric}}} & Object 3DIT &  14.06 & 0.284 & 0.411 & 0.848 & 17.02 & 0.500 & 0.340 & 15.71  \\
& Neural Assets & 13.74 & 0.221 & 0.428 & 0.826 & 16.73 &  0.414 & 0.388 & 23.08 \\
& \ourmethod & 18.90 & 0.557 & 0.227 & 0.914 & 21.32 & 0.674 & 0.198 & 9.11 \\
\midrule
\multirow{3}*{\shortstack[c]{Objectron~\cite{ahmadyan2021objectron}}} & Object 3DIT & 13.88 & 0.290 & 0.424 & 0.902 & 14.98 & 0.355 & 0.423 & 6.14   \\ %
& Neural Assets & 13.73 & 0.278 & 0.427 & 0.921  & 14.56 & 0.337 & 0.427 & 6.18  \\ %

& \ourmethod  & 16.06 & 0.389 & 0.291 & 0.959 & 16.54 & 0.413 & 0.323 & 3.25  \\ %
\midrule
\multirow{3}*{\shortstack[c]{WOD~\cite{sun2020waymo-open}}} & Object 3DIT & 18.90 & 0.448 & 0.255 & 0.930 & 23.21 & 0.640 & 0.220 & 11.92 \\
& Neural Assets &  16.87 & 0.301 & 0.322 & 0.901 & 20.41 & 0.548 & 0.267 & 15.39  \\ 
& \ourmethod &  20.93 & 0.596 & 0.185 & 0.956 & 24.11 & 0.676 & 0.189 & 10.02 \\ 
\bottomrule
\end{tabular}%
} %
\caption{Quantitative results with MOVi-E, Objectron, and Waymo Open Dataset (WOD). The model controls both the objects and the camera to transform the source frame into the target frame of a video.  
Please see \S\ref{exp:setting} for details of the baselines and datasets.
}
\label{tab:main_quantitative}
\end{table}

\begin{figure}[t!]
\centering
\includegraphics[width=0.8\linewidth]{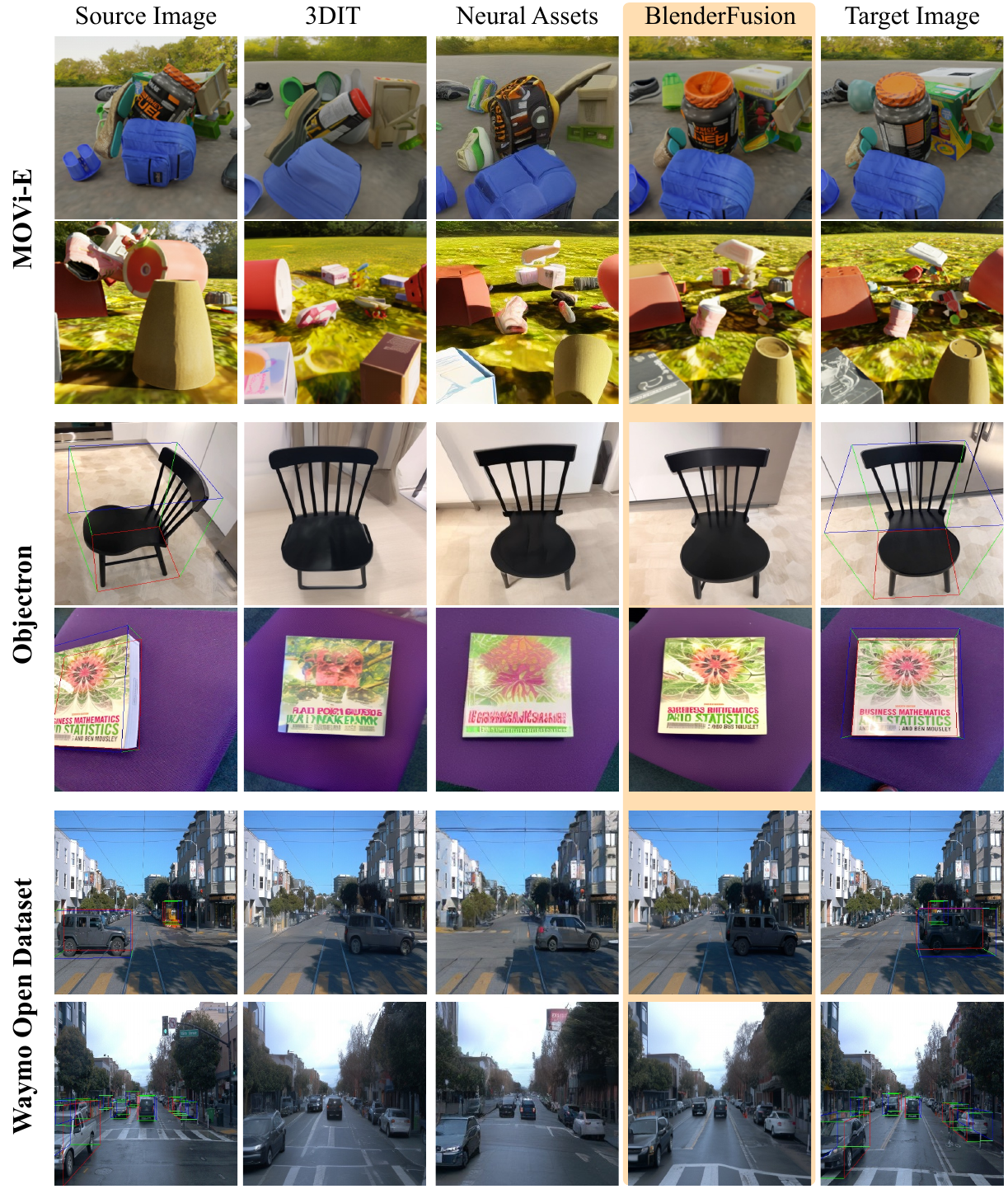}
\caption{Qualitative results with the standard video frame setup on three datasets. \ourmethod outperforms the baselines in visual quality and object identity preservation. The 3D bounding boxes in MOVi-E are omitted for clear visualization.
}
\label{fig:standard_qualitative}
\end{figure}

\subsection{Standard Evaluation}
\label{exp:standard_results}

\autoref{tab:main_quantitative} and 
\autoref{fig:standard_qualitative} summarize our comparisons with baselines in the standard video frame setup. 
Quantitatively, \ourmethod consistently improves both object-level and image-level metrics across all three datasets, indicating better modeling of both foreground and background.
Qualitatively, our method preserves precise object appearance and identity well, while correctly capturing geometry and shadings. In contrast, baselines often distort details (\eg, the geometry of the chair back in the first example of Objectron, the shape of the car in the first example of WOD). Additionally, the baselines struggle to handle a large number of objects with various transforms in the synthetic MOVi-E data with high dynamics, while ours shows reliable generation quality over multiple objects.

However, since this evaluation setup entangles camera and object dynamics, it is unsuitable for evaluating disentangled control capabilities. 
For example, a model that entirely overfits solely to camera motion (\ie, novel-view synthesis) achieve great performance on the standard Objectron evaluation setup. At test time, an important use case is to keep the camera frozen while manipulating the objects and background. Therefore, the next subsection presents further comparisons on tasks requiring more fine-grained and disentangled control capabilities.

\begin{figure}[t!]
\centering
\includegraphics[width=\linewidth]{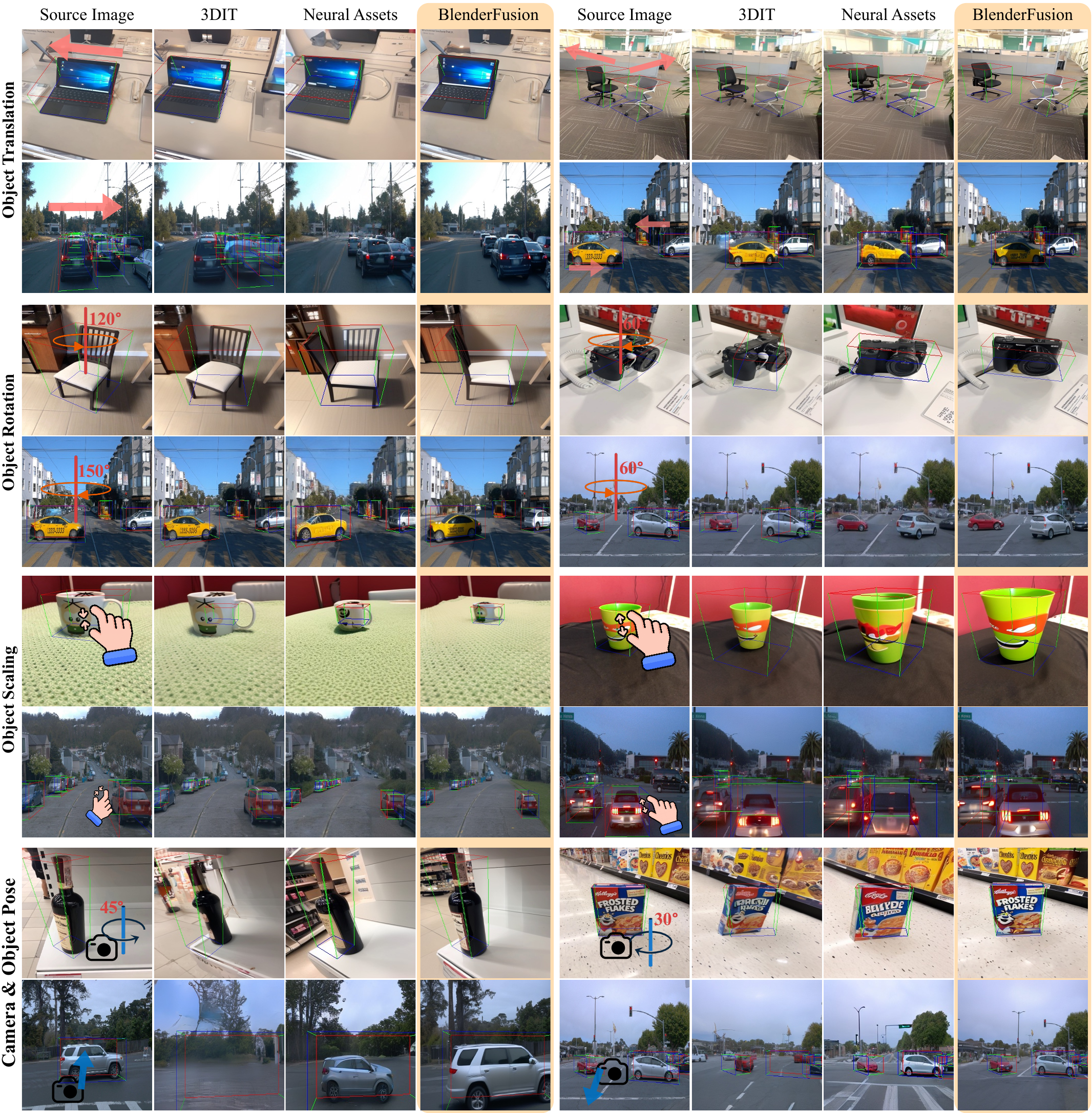}
\caption{Qualitative comparison results on disentangled visual control tasks, using a source image from either Objectron or WOD. \ourmethod demonstrates more precise control, more consistent object identity, and better disentanglement of camera and object.}
\label{fig:prelim_control}
\end{figure}

\subsection{Disentangled Control and Fine-grained Compositing}
\label{exp:finegrained_results}

\mypara{Disentangled Control} \autoref{fig:prelim_control} compares three methods across four types of disentangled control tasks: object translation, rotation, and scaling with a fixed camera, as well as jointly specified camera motion and object pose.
While 3DIT performs reasonably well in the standard video setting, it fails on nearly all disentangled object manipulation tasks and tends to keep the object still. 
This suggests a strong entanglement between object and camera motion.
NA demonstrates significantly better disentangled control than 3DIT but has two major limitations:
1) It loses appearance and geometric details for both objects and backgrounds because its DINO features are from low-resolution images, and this encoder likely discards fine details even after fine-tuning.
2) Foreground and background interfere with each other. This issue arises because RoIAlign on DINO features cannot clearly separate objects from the background.
In contrast, our method precisely follows the intended 3D transformation while decently preserving both geometry and appearance details. Furthermore, when the camera remains fixed, the background remains stable, as Blender can accurately simulate the camera change.

\begin{figure}[t!]
\centering
\includegraphics[width=\linewidth]{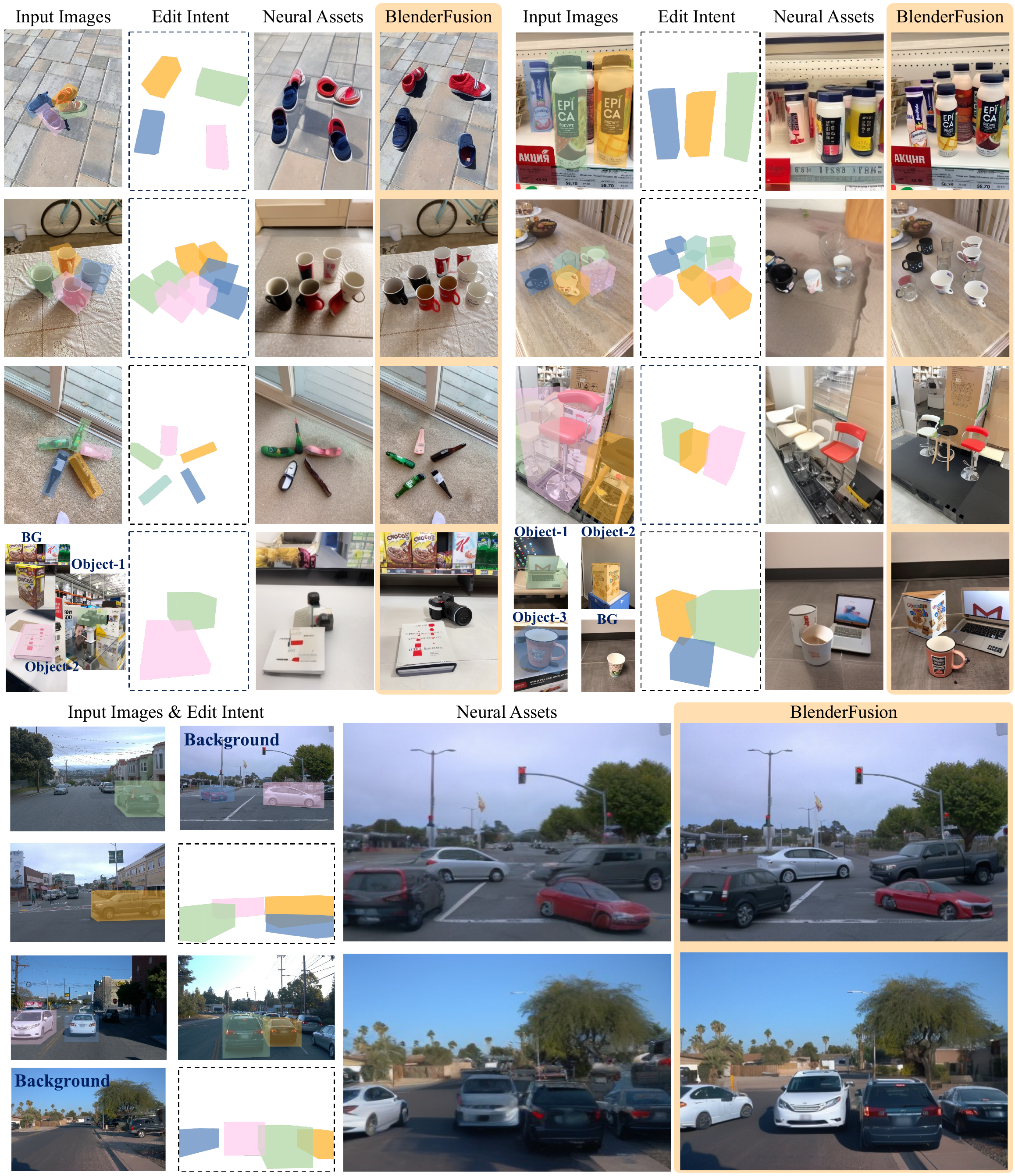}
\caption{
Qualitative results on fine-grained multi-object editing and compositing tasks. The edit intent demonstrates the expected output geometry while the color encodes the object identity.
}
\label{fig:finegrained_control}
\end{figure}

\mypara{Fine-grained Editing and Compositing}
We further explore more complex 3D-aware editing and compositing tasks in \autoref{fig:finegrained_control}, comparing \ourmethod with NA. 
As task complexity increases, the benefits of explicit 3D grounding become more pronounced. Concretely,

\vspace{-4pt} \noindent \raisebox{0.3ex}{\tiny$\bullet$} In the first row, NA misaligns spatial transformations (\eg, positions and poses of shoes and bottles) and mixes appearances, whereas ours ensures both geometric and semantic consistency. NA struggles with close objects due to RoIAlign, while our layering step leverages visual foundation models for precise object delineation and lifting and performs accurate geometric transformations in Blender.

\vspace{-4pt} \noindent \raisebox{0.3ex}{\tiny$\bullet$} In the second row, NA fails to duplicate objects correctly, generating only 5 out of 8 cups. Moreover, the duplicated objects exhibit undesired changes in appearance and shape, whereas we faithfully generate all objects while preserving both appearance and geometry. NA struggles with handling many objects as it falls outside its training distribution, while ours explicitly performs duplication within Blender, bypassing this limitation.

\vspace{-4pt} \noindent \raisebox{0.3ex}{\tiny$\bullet$} In the third row, NA loses original object appearances during object shuffling and swapping (\eg, a black chair disappears) and fails to maintain depth consistency, whereas ours preserves both appearance and natural perspective shifts. 

\vspace{-4pt} \noindent \raisebox{0.3ex}{\tiny$\bullet$} In the fourth row, NA produces inaccurate poses and disregards shading, whereas ours ensures both geometric accuracy and lighting consistency.

\vspace{-4pt} \noindent \raisebox{0.3ex}{\tiny$\bullet$} In the WOD multi-image recomposition tasks from the last two rows, NA produces acceptable results but still loses a lot of object details due to the highly lossy DINO encoding. Ours preserves better object identity and fidelity, while showing more coherent shadings.

These results demonstrate the impact of our core idea of decoupling control from generation, leveraging Blender as a bridge to achieve precise and flexible visual compositing.

\subsection{More Results}
\label{exp:more_results}

\phantom{.}\begin{wraptable}[8]{r}{0.5\textwidth}
\centering
\small
\vspace{-3em}
\resizebox{\linewidth}{!}
{
\begin{tabular}{@{}lccc@{}}
\toprule
\textbf{Setting} & \textbf{Ours (\%)} & \textbf{Baseline (\%)} & \textbf{Draw (\%)} \\
\midrule
Overall       & 87.04 & 6.40 & 6.56 \\
\midrule
Video & 80.79 & 8.80 & 10.42 \\
Disentangled & 88.37 & 6.60 & 5.03 \\
Fine-grained & 93.75 & 2.43 & 3.82 \\
\bottomrule
\end{tabular}
}
\caption{Human evaluation results for three setups.}
\label{tab:human_eval}
\end{wraptable}

\vspace{-3.5em}
\mypara{Human Evaluation}
Based on the three evaluation settings in \S\ref{exp:standard_results} and \S\ref{exp:finegrained_results}, we conduct a user study to further validate the superority of our method. 
We curate 54 examples for human evaluation: 18 for the standard video frame setup, 24 for disentangled object control, and 12 for complex fine-grained compositional control. Participants compared two shuffled generations (ours vs. Neural Assets) and selected one: A, B, or Similar. The results were collected from 1,294 selections from 24 users and are summarized in \autoref{tab:human_eval}. The gap between \ourmethod and the baseline gets larger when the editing task becomes more challenging, which further demonstrates the advantages of the 3D-grounded framework in complex and compositional visual editing.

\begin{figure}[t!]
\centering
\includegraphics[width=\linewidth]{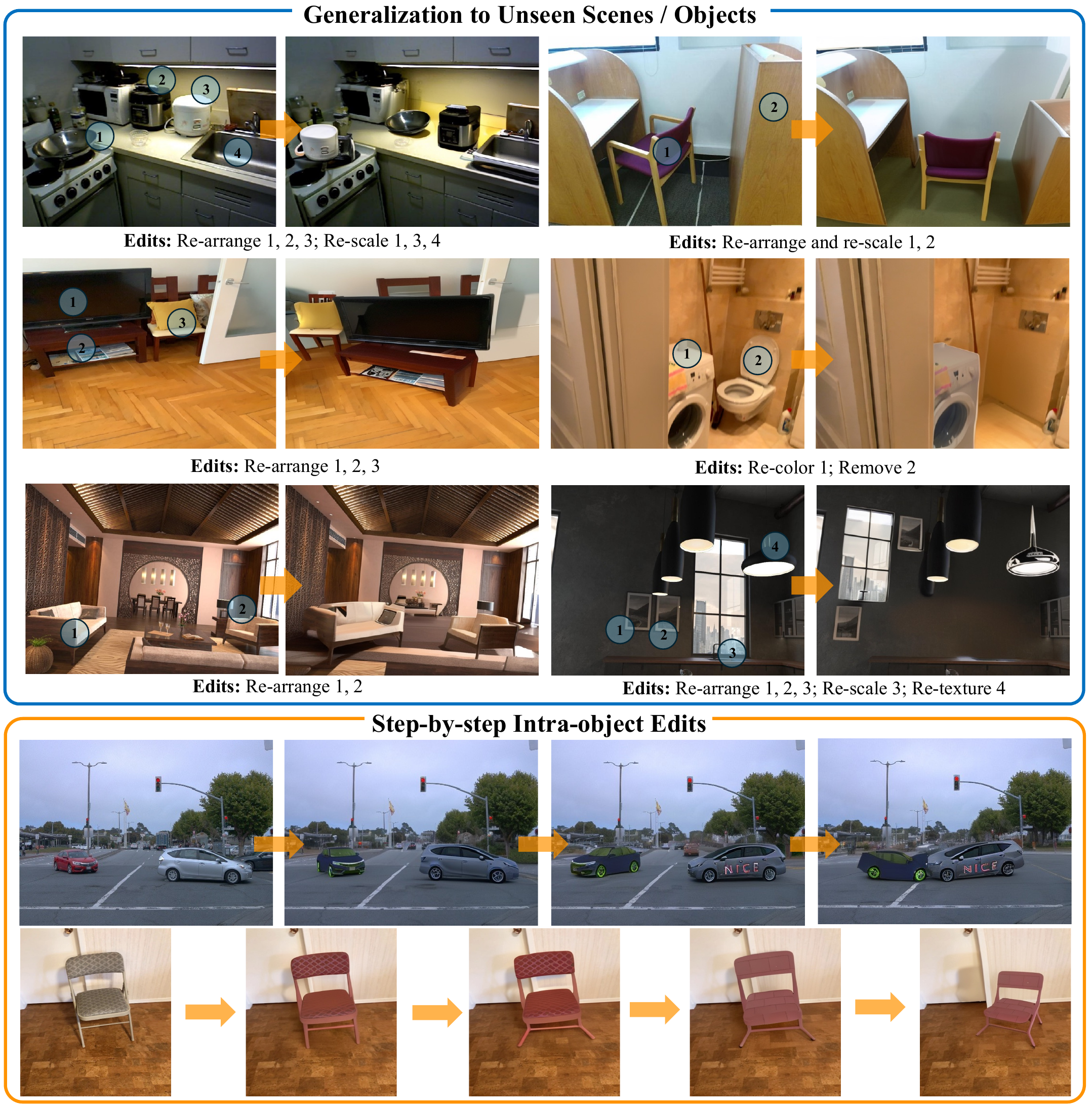}
\caption{\textbf{Top:} \ourmethod shows reasonable generalization to in-the-wild images from SUN-RGBD~\cite{song2015sun-rgbd}, ARKitScenes~\cite{baruch2021arkitscenes}, and Hypersim~\cite{roberts2021hypersim} datasets. \textbf{Bottom:} our framework inherits the versatile editing capabilities of graphics software, enabling diverse object control tasks beyond the training data. Images are resized to facilitate visualization.
}
\label{fig:more_results}
\end{figure}

\mypara{Generalization}
We apply \ourmethod trained on Objectron data to in-the-wild images from SUN-RGBD~\cite{song2015sun-rgbd}, ARKitScenes~\cite{baruch2021arkitscenes}, and Hypersim~\cite{roberts2021hypersim} datasets. All three datasets present scenes with much higher complexity and richer details than Objectron. \autoref{fig:more_results} (Top) demonstrates the generalization results. Those in-the-wild images present much more complicated scene structure and object details than the Objectron training data, and Hypersim is a high-end synthetic dataset created by professional designers. \ourmethod presents reasonable generalization capabilities, although the visual quality shows some degradation compared to in-domain results.

\mypara{Progressive Editing} 
In \autoref{fig:more_results} (Bottom), we demonstrate progressive, step-by-step editing, where each intermediate modification in Blender is rendered using our diffusion compositor. For these examples, Hunyuan3D v2 is utilized during the layering step to generate higher-quality 3D object models, as detailed in \S\ref{method:pipeline}.
The first row illustrates: 1) color changes for each object, 2) rotation and text decal application, and 3) object deformation.
The second row shows: 1) a color change, 2) part-level deformation, 3) texture replacement, and 4) rotation.
Although we showcase simple, progressive edits here, more advanced controls supported by Blender are also possible.

\begin{figure}[t!]
\centering
\includegraphics[width=0.8\linewidth]{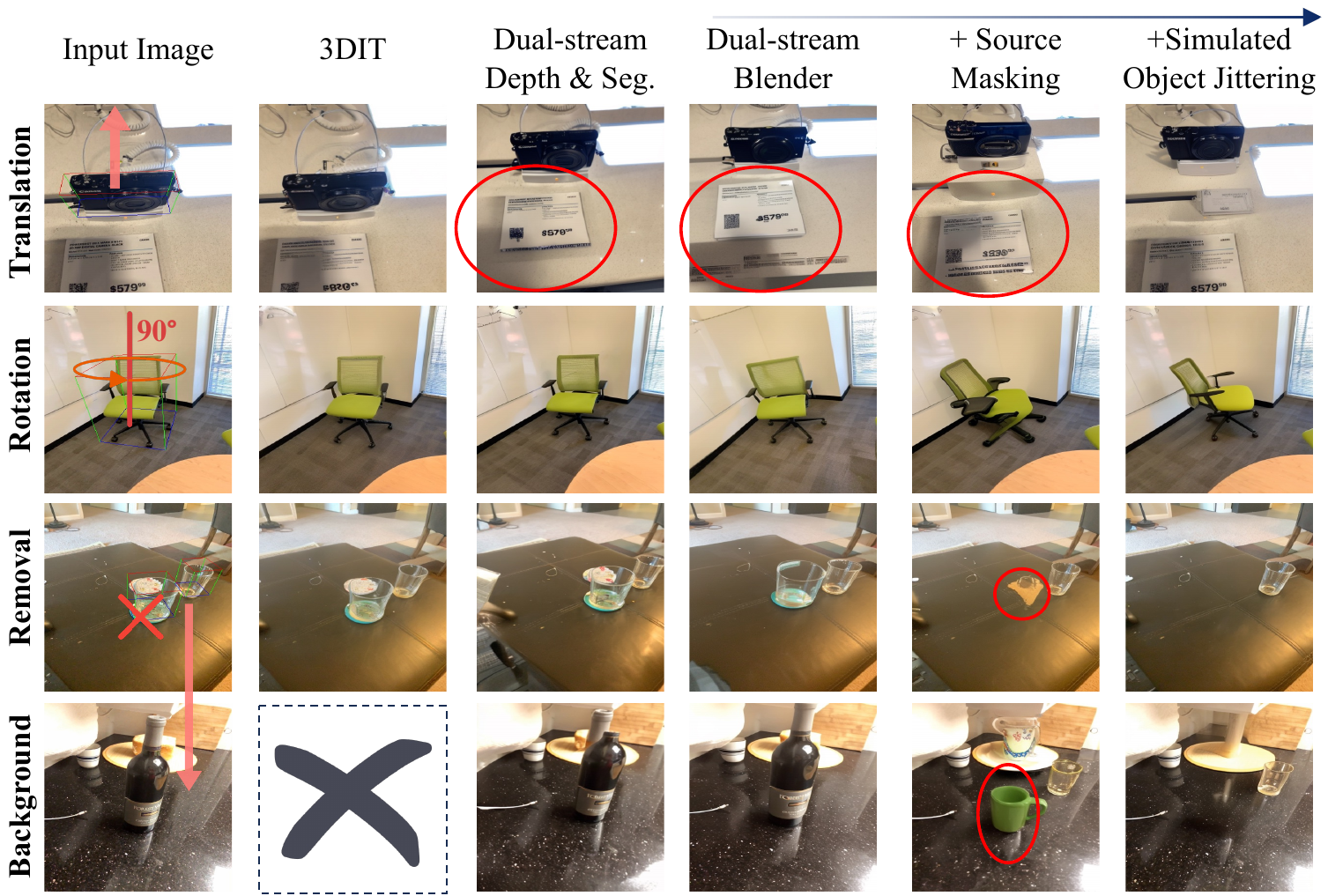}
\caption{Qualitative ablation results over the key designs, including using Blender renders as the 3D-grounded control signals as well as the training strategies of the diffusion compositor.}
\label{fig:ablation}
\end{figure}

\mypara{Ablation Study}
As discussed earlier, the standard video frame setup does not indicate disentangled control capabilities. Thus, we focus on qualitative analysis for ablation studies here and leave the quantitative results to the \supp. 
\autoref{fig:ablation} investigates the impact of three key design choices: 1) the dual-stream architecture, 2) the source masking training strategy, and 3) the simulated object jittering training strategy. 
A straightforward ablation study is to use our dual-stream design while adding the depth and segmentation of the input image from Depth Pro and SAM2 into the source stream (third column). 
However, this model only shows improvements over the 3DIT baseline in the object translation setting.
When adding the Blender renders, the dual-stream model performs reasonable object translation (first row), but the background still consistently moves along with the object. This variant also struggles with disentangled object rotation (second row) and instance-wise removal (third row), and fails in background changes (fourth row). 
Introducing the source masking strategy clearly strengthens object-level control, but the model still suffers from entangled camera and object states. 
This is mainly because the training data rarely covers object manipulation under a fixed camera.
The simulated object jittering strategy addresses these remaining issues, significantly enhancing the disentangled control capabilities of both object and background. 
Note that the source masking and simulated object jittering are designed based on the Blender renders -- without the $\render^\source$ and $\render^\target$ in the compositor's inputs, both training strategies become invalid.

\section{Conclusion}
This paper introduces \ourmethod, a novel framework that integrates generative diffusion models with 3D graphics tools, enabling fine-grained and highly controllable visual compositing.
\ourmethod follows three key steps of traditional visual compositing: it segments and lifts 2D images into editable 3D entities, performs precise manipulations within Blender, and refines the resulting visuals through a specialized generative compositor. By leveraging this pipeline, we achieve superior control and realism compared to prior techniques.
Experiments on real-world video datasets demonstrate that \ourmethod significantly advances multi-object scene editing, providing a flexible and practical solution for complex visual content creation.

\vspace{-0.2cm}
\section*{Acknowledgements}
\vspace{-0.2cm}
We thank Ziyi Wu for discussions on Neural Assets and synthetic data pre-training. We thank Thomas Kipf and Caroline Pantofaru for reviewing the paper and providing feedback.

\bibliographystyle{abbrvnat}
\nobibliography*
\bibliography{references}

\newpage 

\appendix 
\part*{Appendices}

\noindent The appendix provides remaining implementation details and additional experimental results as mentioned in the main paper:

\begin{itemize}[leftmargin=*,itemsep=1pt]

\item[$\diamond$] \S\ref{supp:details} presents  the remaining implementation details of \ourmethod, and describes full details about the baselines and datasets. 

\item[$\diamond$] \S\ref{supp:results} provides more qualitative results in a similar style of \autoref{fig:standard_qualitative} and \autoref{fig:prelim_control} of the main paper, the quantitative ablation results corresponding to \autoref{fig:ablation}, and failure case analyses.

\end{itemize}

\section{Remaining Implementation Details}
\label{supp:details}

\subsection{Remaining Details of \ourmethod}
\label{supp:method_details}

\vspace{-0.3em}\mypara{Source Masking Details}
During training, the source masking is always consistently applied to the source image and the source Blender render. Concretely, we randomly mask each foreground object with an (independent) probability of 0.5, then apply random masking to the background using boxes with a similar aspect ratio as the corresponding foreground objects. The mask is applied by masking out the 2D bounding box derived from the 3D object bounding box, with a dilation operation. The background masking here effectively alleviates the object inpainting bias, which might hurt the performance in object removal. When an object is masked out, the corresponding source bounding box information is also dropped for the source stream. The simulated object jittering training also applies the same source masking strategy, while the masks applied to the ``source'' image and source render are different -- because the ``source'' image in this reconstruction setting is actually the target frame, while the source render still comes from the true source frame.

\vspace{-0.3em}\mypara{Frame Sampling Strategy for Training Data}
To train \ourmethod, we need to prepare the Blender renderings of the source and target frames. For both datasets, we sample the source frame uniformly with a fixed stride. Then, for each source stream, we randomly sample a set of target frames. For each pair of source and target frames, we obtain the object 3D models with the 3D lifting process, import them into Blender, and simulate the object and camera transforms. In this way, we obtain the source and target renders for training our dual-stream diffusion model.
Note that since the two baselines do not rely on the reconstruction and re-render steps, there is no limitation on the sampling of the source and target frames, and they actually train with more diverse data than \ourmethod. The strong quantitative and qualitative performance of our method suggests that training with stronger 3D grounding can also improve the data efficiency.

\vspace{-0.3em}\mypara{Test-time Details}
For the standard video evaluation setting, all conditions are directly passed to the model without any masking or dropping. For disentangled object manipulation with fixed camera (including translation, rotation, and scaling), the source masking is applied to two regions in the \textit{source image}: 1) the region of the original object and 2) the expected region of the target object. The \textit{source render} and the \textit{source object 3D bounding box} are only masked/dropped if the object is intended to be removed or replaced, otherwise, they are preserved and will be used as the main reference for object appearance and geometry. The source masking at test time encourages better disentanglement between foreground objects and background contexts.

When applying the advanced object-level control inherited from Blender (\eg, attribute change, deformation), as described in \S\ref{method:pipeline}, those edits are always reflected in both the source render $\render^\source$ and target render $\render^\target$ in the dual-stream inputs. Concretely, when we close the screen of the laptop like in \autoref{fig:teaser}, the edit is always first reflected in the initial scene $\scene^\source$ and then rendered to update $\render^\source$, then, further edits like object rotations or translations transform $\scene^\source$ into $\scene^\target$ and we render  $\scene^\target$ to get $\render^\target$.  In this way, the difference between the target and source stream is always the basic object transformations, which is consistent with the training setting and makes the framework generalizes well to much more complex multi-object editing and scene composition tasks at test time.

\subsection{Remaining Details of Datasets and Baselines}
\label{supp:detail_data_and_baseline}

\mypara{MOVi-E Details}
Instead of using the original 10K MOVi-E videos released by the Kubric dataset generator~\cite{greff2022kubric}, we update the generation config to generate a new set of 10K videos with more extensive object and camera motions. Specifically, we update the number of static objects from 10-20 to 5-10, and the number of dynamic objects from 1-3 to 5-10. For the camera movement, we reset the maximum camera movement from 4 units in Kubric's 3D space to 8 units. This new variant is more suitable for evaluating the model 3D controllability over multiple objects and the camera. Additionally, it makes the synthetic MOVi-E data a more effective pre-training resource for real-world datasets with limited object and camera motions (\eg, WOD has highly imbalanced object poses).

\vspace{-0.3em}\mypara{Objectron Details}
Simialr to NA's setting, we drop the bike category and use the remaining data for training and evaluation. For each video, we randomly sample 60 frames without repeating to get a clip. Then, we randomly sample source and target frames for each clip. The test set contains 2812 videos, and we sample 67,092 pairs for evaluation. The original resolution of the dataset is 480$\times$640 (height is larger than the width). We keep the aspect ratio and use 384$\times$512.

\vspace{-0.3em}\mypara{Waymo Open Dataset (WOD) Details}
Similar to the settings of NA, we only take the front-view camera of WOD, and use the same filtering strategies to filter out the videos without large objects. Concretely, the pair of source and target frames is considered as valid only when there is at least one large object that occupies more than 1\% image area in both frames. We sample 6976 pairs of source and target frames from the test set for evaluation.

\vspace{-0.3em}\mypara{Baseline Implementation Details}
As described in \S\ref{exp:setting}, we re-implement the two major baselines, 3DIT and Neural Assets (NA),  and carefully control the experimental settings to focus on the essential difference between the model designs for object/scene controllability.
The re-implemented 3DIT uses the same base model and image resolution as ours, and shares exactly the same training recipes and inference setups. It can be considered as ours without the two-stream model architecture, the grounded Blender renders, and the two accompanying training strategies.
In our experiments, we observe that training NA with high generation resolutions (\ie, 384$\times$512 for Objectron, 528$\times$352 for WOD) does not bring improvement on quantitative metrics, but produces apparent object appearance shifts compared to the original 256$\times$256 setting. This might be related to the resolution inconsistency between its DINO encoder (\ie, 224$\times$224) and diffusion model, and is not trivially resolved by jointly fine-tuning all modules. Increasing DINO's input resolution would significantly raise the training GPU memory cost and make the training unaffordable, which also has the risk of invalidating the pre-trained priors. Therefore, for better qualitative results under DINO's pre-defined input resolution, we choose to keep NA's original 256$\times$256 generation resolution in all experiments.

\section{Additional Experimental Results}
\label{supp:results}

\subsection{Quantitative Ablation Study}

\autoref{tab:ablations} provides the quantitative ablation evaluation results with the standard video setting. The dual-stream design alone can improve the baseline significantly by more effectively leveraging the source and target information. Incorporating the Blender renders in both streams further unlocks the model's capacity by providing reliable 3D groundings.
However, as presented in \autoref{fig:ablation}, these variants still cannot achieve disentangled object control, struggling to manipulate objects with a fixed camera or to do significant modifications to the source image (\eg, object removal, background change). \begin{wraptable}[12]{r}{0.5\textwidth}
\centering
\scriptsize
\resizebox{\linewidth}{!}{
\begin{tabular}{@{}lcccc@{}}
\toprule
\setlength{\tabcolsep}{2.5pt}
\multirow{2}{*}{Method} & \multicolumn{3}{c}{Object-level Metrics} \\ 
\cmidrule(r){2-4}
 &  PSNR$\uparrow$ & SSIM $\uparrow$ & LPIPS $\downarrow$ & FID $\downarrow$ \\
\midrule
3DIT (one-stream) & 13.88 & 0.290 & 0.424  & 6.14  \\ 
\hdashline
Dual-stream (DS) & 15.90 & 0.378  & 0.310  & 3.52 \\
DS + Depth \& Seg. & 16.04 & 0.382 & 0.313  & 3.74 \\
\hdashline
DS + Blender & 16.05 & 0.389  & 0.292   & 2.93 \\ 
+ Source Masking  & 16.18 & 0.393  & 0.290  & 2.64  \\
+ Sim. Obj Jittering  & 16.06 & 0.389  & 0.291 & 3.25 \\
\bottomrule
\end{tabular}
}
\caption{The quantitative ablation studies on the key elements of our diffusion model compositor, corresponding to \autoref{fig:ablation} of the main paper.}
\label{tab:ablations}
\end{wraptable}

The source masking can slightly improve the quantitative results, as it can be considered as a data augmentation strategy to the basic video training setting. 
While being the core design to acquire disentangled object control, the simulated object jittering training strategy slightly lowers the quantitative results on the standard video setting. This is expected because it is essentially an image reconstruction training, and has a different source stream setting from the standard video setup (\ie, source camera is the same as the target camera).

\subsection{Failure Cases}
\label{supp:failure}

\begin{wrapfigure}[12]{r}{0.5\textwidth}
\centering
\includegraphics[width=\linewidth]{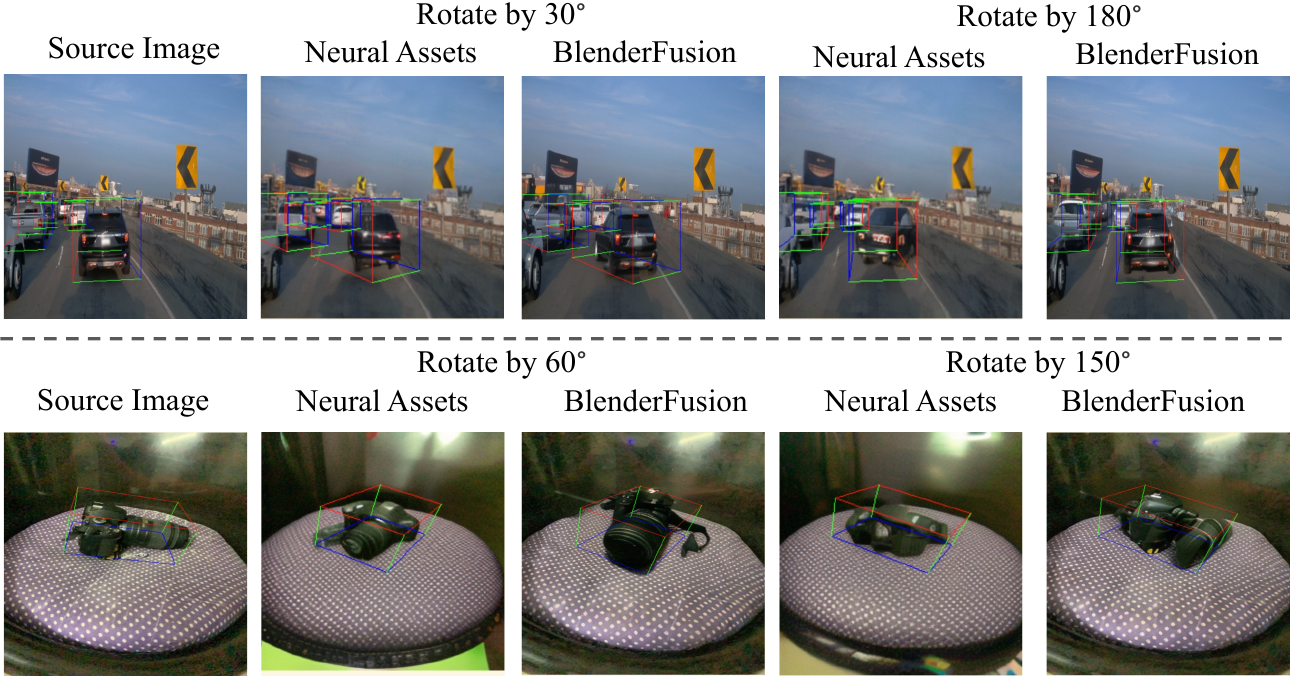}
\caption{Failure cases on object rotation.}
\label{fig:failure_case}
\end{wrapfigure}

In our experiments, one main failure mode is that the model sometimes has difficulty achieving accurate manipulation for disentangled object rotation (\autoref{fig:failure_case}). On WOD, this is largely alleviated by using the MOVi-E pre-trained model as the initialization to compensate for the lack of pose difference between the source and target frames in the training data (as most cars keep going straight). The top row of \autoref{fig:failure_case} shows typical failure cases of NA and ours when not using MOVI-E pre-training. 

On Objectron, the failure of object rotation usually comes with wrong or obviously altered object geometry, and is caused by two factors: 1) the model lacks accurate 3D understanding for complex object geometry (\eg, high-end cameras), and 2) when the object reconstruction is 2.5D, the renders can be unreliable when the object is rotated significantly. For the latter case, using 3D-Gen meshes resolves the problem in most cases, albeit at the cost of a longer processing time for running an image-to-3D model and an object pose alignment step. Pre-training on MOVi-E does not help with this failure case, probably because the objects in MOVi-E only present simple geometry. Two potential directions can be explored in future works: 1) pre-training on datasets with diverse object geometries and motions, or 2) extending the pipeline to multi-view or video input so that a more complete 3D reconstruction can be obtained with state-of-the-art multi-view reconstruction algorithms.

\begin{figure*}[ht]
    \centering
    \includegraphics[width=0.95\linewidth]{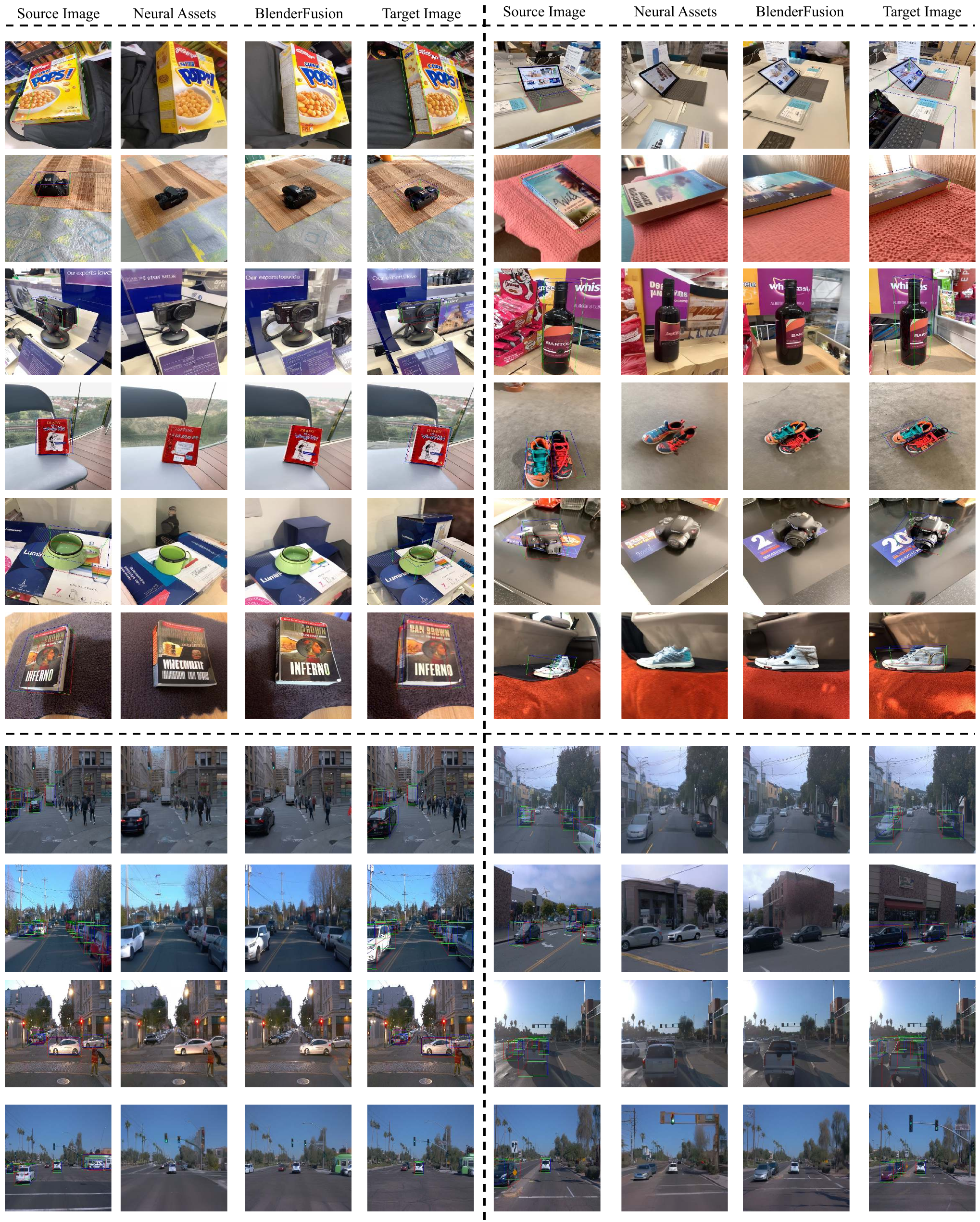}
    \caption{Additional qualitative results of the standard evaluation setting (source and target frames from a video), extending \autoref{fig:standard_qualitative} of the main paper. 3DIT is omitted in this figure.}
    \label{fig:supp:standard_qualitative}
\end{figure*}

\begin{figure*}[ht]
    \centering
    \includegraphics[width=0.95\linewidth]{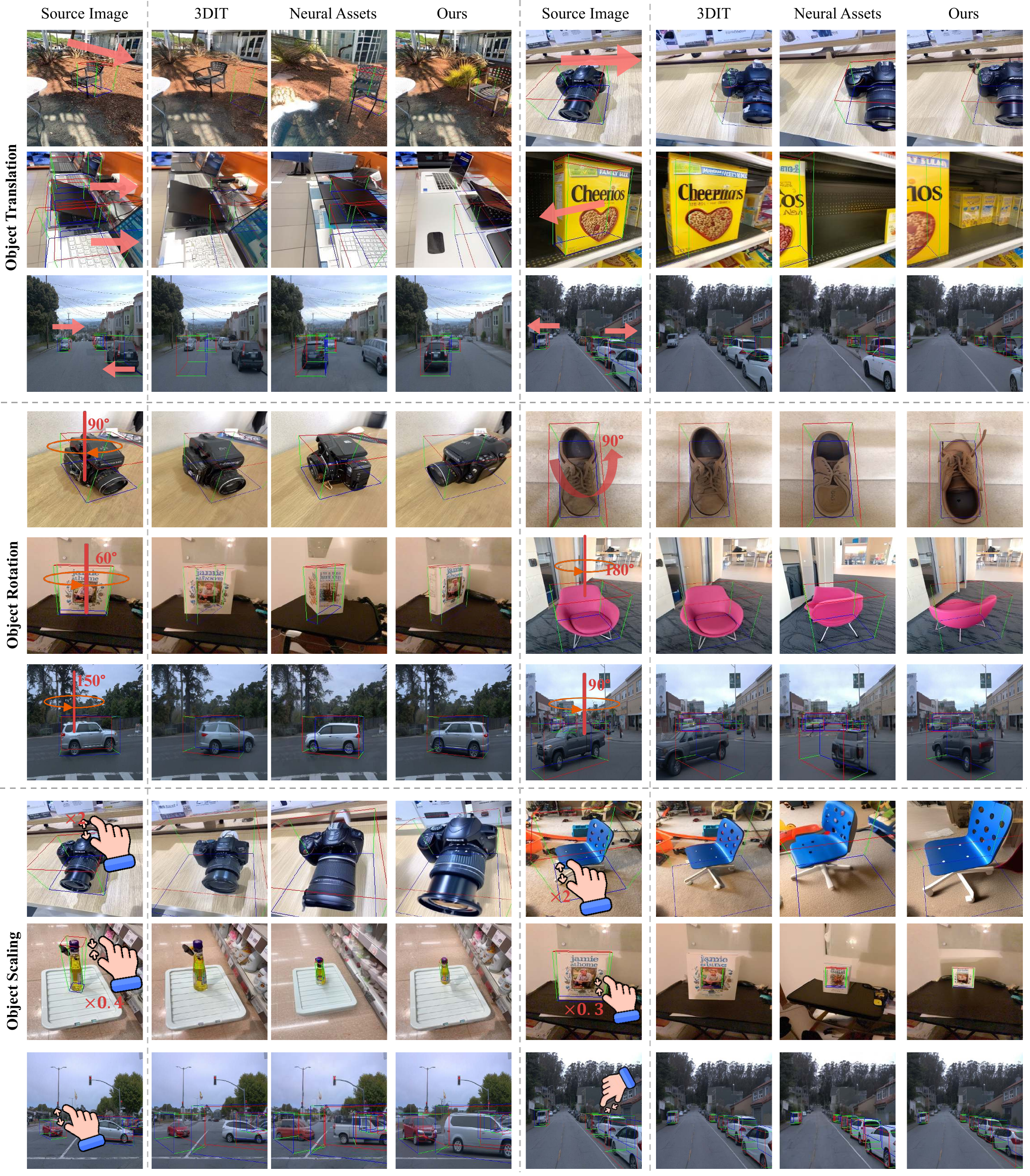}
    \caption{Additional qualitative results of disentangled object control with fixed camera, extending \autoref{fig:prelim_control} of the main paper.}
    \label{fig:supp:prelim_control}
\end{figure*}

\subsection{Additional Qualitative Results}

\autoref{fig:supp:standard_qualitative} and \autoref{fig:supp:prelim_control} provide additional qualitative comparison, extending \autoref{fig:standard_qualitative} and \autoref{fig:prelim_control} of the main paper. These examples further demonstrate the superior object control capability of \ourmethod.

\end{document}